# Quantifying Uncertainty in Deep Learning Classification with Noise in Discrete Inputs for Risk-Based Decision Making

Maryam Kheirandish[a], Shengfan Zhang[a], Donald G. Catanzaro[b], Valeriu Crudu[c]

[a] *Department of Industrial Engineering, University of Arkansas, Fayetteville, AR, USA*

[b] *Department of Biological Sciences, University of Arkansas, Fayetteville, AR, USA*

[c] *Institute of Phthisiopneumology "Chrirl Draganiuc", Chisinau, Moldova*

[c] *State University of Medicine and Pharmacy "Nicolae Testemitanu", Chisinau, Moldova*

**Abstract**

The use of Deep Neural Network (DNN) models in risk-based decision-making has attracted extensive attention with broad applications in medical, finance, manufacturing, and quality control. To mitigate prediction-related risks in decision making, prediction confidence or uncertainty should be assessed alongside the overall performance of algorithms. Recent studies on Bayesian deep learning helps quantify prediction uncertainty arises from input noises and model parameters. However, the normality assumption of input noise in these models limits their applicability to problems involving categorical and discrete feature variables in tabular datasets. In this paper, we propose a mathematical framework to quantify prediction uncertainty for DNN models. The prediction uncertainty arises from errors in predictors that follow some known finite discrete distribution. We then conducted a case study using the framework to predict treatment outcome for tuberculosis patients during their course of treatment. The results demonstrate under a certain level of risk, we can identify risk-sensitive cases, which are prone to be misclassified due to error in predictors. Comparing to the Monte Carlo dropout method, our proposed framework is more aware of misclassification cases. Our proposed framework for uncertainty quantification in deep learning can support risk-based decision making in applications when discrete errors in predictors are present.

# 1. Introduction

Decision-making based on predictions, especially in classification problems, has been a subject of interest since machine-learning (ML) models were invented. To understand and manage the risk of using predictions in decision making, it is essential to quantify the uncertainty of each prediction in addition to assessing the overall performance of these learning algorithms. Quantifying the uncertainty associated with such algorithms is not trivial, and the complexity of calculation increases as the nonlinearity of such algorithms increases. In this essence, uncertainty quantification (UQ) of Deep Neural Network (DNN) classification models, which have a nonlinear and complex mathematical structure, is especially challenging. Despite this challenge, DNN has become a powerful prediction tool in many fields from language processing, finance, investment, engineering to medicine. Although the application of these models has been tied to image and video processing for years, they are becoming more popular for tabular data with various applications such as to predict disease outcomes (De Angeli et al. 2022), credit risk (Chang et al. 2022), exchange rate (Mourtas et al. 2023), and vehicle emissions (Seo and Park 2023). The growing application of DNN classification models in life-threatening problems, such as medical decision making, increases the importance of assessing the prediction uncertainty of these models.

## 1.1 Uncertainty Definition

Adopting the general definition of uncertainty as any deviation from complete determinism (Walker et al. 2003), several uncertainty typologies have been created for various purposes. Suppose $D$ is the dataset consisting of $N$ training data points, $\{(\mathbf{x}_1, y_1), (\mathbf{x}_2, y_2), \ldots, (\mathbf{x}_N, y_N)\}$ where $y_i = \{1, 2, \ldots, C\}$ for $i = 1, \ldots, N$. For each new query $x^{pred}$, the classification model outputs predictive distribution $p_c(x^{pred}) =$ $p(y^{pred} = c|x^{pred}, D)$. The predicted class is then $c^* = \underset{1,2,...,C}{\operatorname{argmax}} p_c(x^{pred})$. There are two common metrics in the literature for quantifying the uncertainty of a prediction, variation ratio and predictive entropy (Gal 2016). The variation ratio approximates the quantity $1 - p_{c^*}(x^{pred})$, whereas predictive entropy quantifies the degree of uncertainty for the predictive distribution, $p_c(x^{pred})$.

*1.2. Types of Uncertainty*

Our research is focused on the uncertainty that should be considered in model-based decision-support. Uncertainty can be explored in three dimensions, location, level, and nature according to the literature on uncertainty management (Walker et al. 2003). The location of uncertainty identifies where it is demonstrated within the whole model, which can be within the context, model structure, inputs, parameters, or outcomes. There are four levels of uncertainty to recognize when developing strategies to cope with uncertainty in policy making, i.e., statistical uncertainty, scenario uncertainty, recognized ignorance, and total ignorance or deep uncertainty (Walker et al. 2003). Based on this definition, recognized ignorance can further be divided into reducible and irreducible ignorance. The nature of uncertainty, which can be epistemic or aleatoric, is the dimension of uncertainty that has been welcomed by the machine learning community in recent years. Epistemic uncertainty is due to the lack of knowledge about the behavior of system, and aleatoric uncertainty refers to the notion of randomness (Hüllermeier and Waegeman 2021, Hora 1996). According to this definition, aleatoric uncertainty is irreducible, as it occurs even if the best model is known. In contrast, epistemic uncertainty can be reduced by employing a more proper model. However, aleatoric and epistemic uncertainty should not be considered as absolute notions, since they may turn into each other during the study, based on the context and application of the model (Hüllermeier and Waegeman 2021, Der Kiureghian and Ditlevsen 2009).

Both aleatoric and epistemic uncertainty can occur in any location of the model, from input data to model structure (Walker et al. 2003). To illustrate, consider the binary classification problem represented in Figure 1, in which a model is trained on trainset $D$ using $x_1$ and $x_2$ as predictors with positive and negative labels. Suppose two support vector machines with radial basis function kernel but different hyperparameter initialization are trained to predict labels. Three new queries with true negative label $X^1$, $X^2$ and $X^3$ are given to the model to be classified. Then the prediction has high epistemic uncertainty for $X^1$ due to lack of knowledge, high aleatoric uncertainty for $X^2$ due to stochasticity of data, and high epistemic uncertainty for $X^3$ due to selection of model hyperparameters.

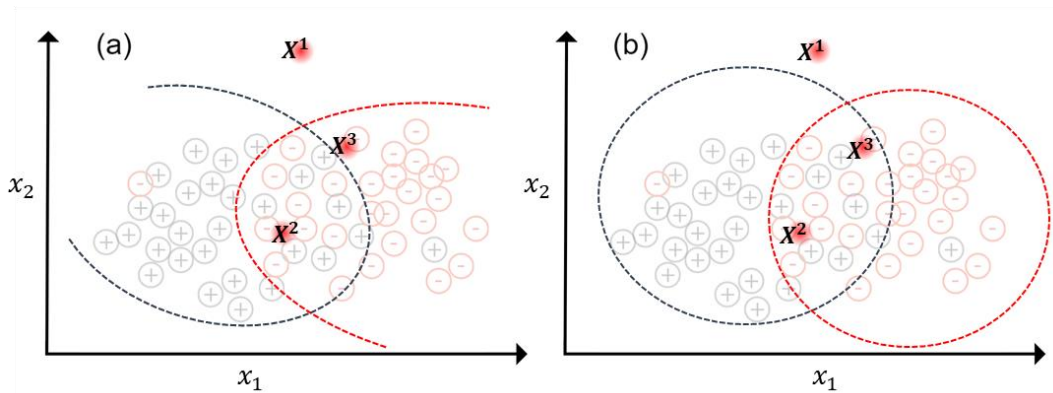

Figure 1: A Binary classification model is trained with two different hypotheses on hyperparameters in (a) and (b). Three new queries $X^1$, $X^2$, and $X^3$ are given to both models. Suppose that all three have true negative labels. The epistemic uncertainty of prediction is high due to lack of knowledge for $X^1$. In this case, collecting more training data may result in a more certain prediction model in favor of this point. The aleatoric uncertainty is high when predicting label for $X^2$ due to inherent randomness of data that is intuitively irreducible. Epistemic

prediction uncertainty is high for $X^3$ since different hypotheses about model hyperparameters will change predictions for this query. Obtaining more knowledge about the behavior of data reduces prediction uncertainty in favor of this point.

In a problem setting, aleatoric uncertainty is often irreducible, meaning that even collecting more training data or information about the behavior of the system may not decrease the inherent randomness of data (Hüllermeier and Waegeman 2021). Although aleatoric uncertainty can be reduced by adding more variables to the model as illustrated in Figure 2, it also changes the problem setting since it changes the set of predictors. Conversely, epistemic uncertainty is known as reducible uncertainty. It implies that collecting more information about data or behavior of the system may decrease this type of uncertainty without any modification to the problem setting.

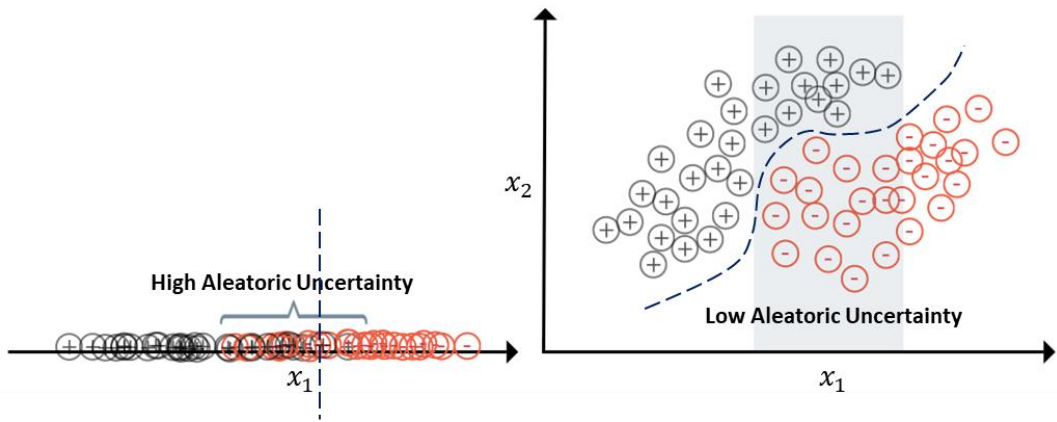

Figure 2: Adding more variables to prediction models can result in decreasing the aleatoric uncertainty, but it modifies the problem setting.

Figure 3(b) illustrates the regions characterized by elevated levels of aleatoric and epistemic uncertainty during prediction when the uncertainty in $X_1$ is not considered. In this context, instances where the new query falls outside the established classification regions result in increased epistemic uncertainty, attributed to the lack of available information. When the new query falls within the intersection of classification boundaries, the aleatoric uncertainty is amplified, stemming from the inherent randomness within the data. Suppose the errors in variable $x_1$ follow a discrete distribution. When observing $x_1$ as $x^*$, the true value can be $\check{x} < x^*$ with a probability of $p_1$, or $\hat{x} > x^*$ with a probability of $p_2$.

Figure 3(c) demonstrates the alterations in predictions (referred to as transition points in the following) and associated uncertainties when the true value of $x_1$ deviates from $x^*$. More specifically, transition 2 highlights the genuine significance of $x_1$ as $\hat{x}$ for the observed value $x^*$, leading to prediction being altered with a probability of $p_2$. In the same manner, transition 6 alters the degree of aleatoric uncertainty with a probability of $p_1$, and transition 4 alters the degree of epistemic uncertainty with a probability of $p_2$. In contrast, Figure 3(d) exemplifies instances where deviations from the observed values of $x_1$ do not impact either prediction or the associated degrees of uncertainty.

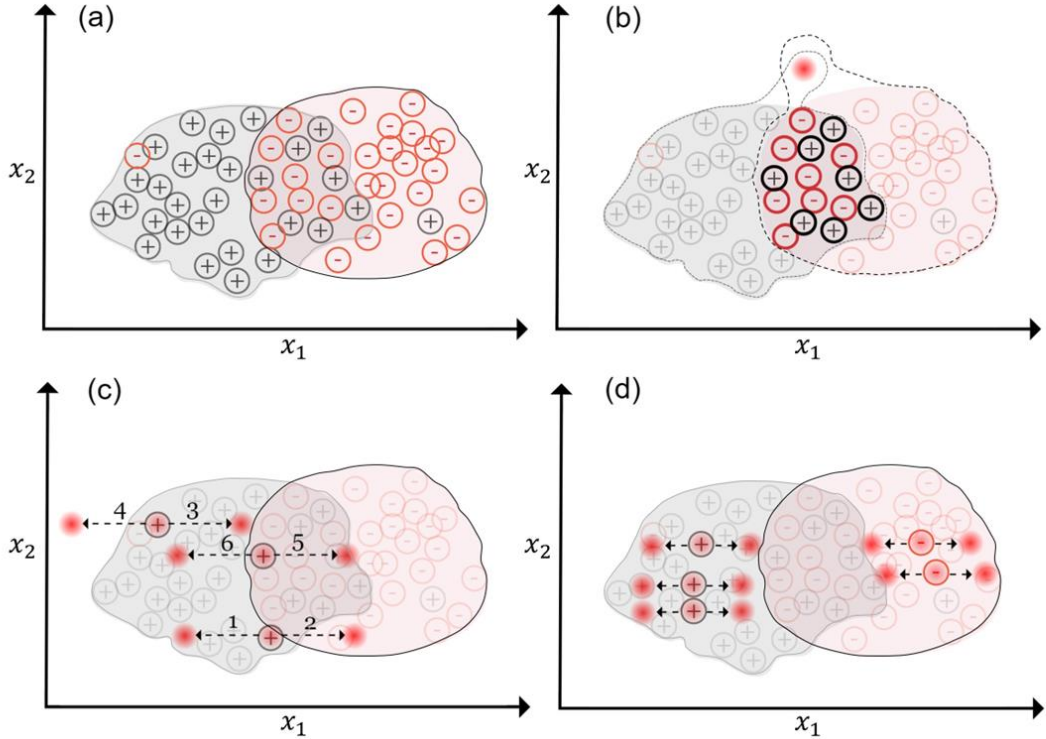

Figure 3: a) Binary classification model trained on dataset $D$ disregarding errors in variables. b) Ignoring the errors in variables, where the new query falls outside the classification regions, results in increased epistemic uncertainty. When the new query falls within the intersection of classification boundaries, the aleatoric uncertainty is amplified, stemming from the inherent randomness within the data. c) Considering errors in variable $x_1$, transition 2 affects the final prediction, transition 6 affects degree of aleatoric uncertainty, and transition 4 increases epistemic uncertainty. d) Considering errors in variable $x_1$, for some instances, and some small deviations from observed values, neither prediction nor corresponding uncertainties are affected.

## *1.3 An illustrative Example in Health Care*

Tuberculosis (TB) is a global health challenge and the number one cause of death due to infectious diseases. Clinical tools aiding physicians in selecting treatment options for drug-susceptible TB are typically limited to one standard anti-TB regime with any changes to regimes driven by reacting to patient side-effects. Effective patient outcome prediction has the potential to assist physicians in developing individualized patient medical treatment course based on patient-based parameters that indicate their progression through therapy. This example addresses the problem of predicting TB treatment binary outcome (i.e., a patient is cured or not) using follow-up information such as laboratory test results in addition to baseline data such as age, gender, and TB type (Kheirandish et al. 2022). Typically, four standard laboratory tests are used to tract patient-progress follow-up are the Acid-Fast Bacillus (AFB) smear, culture, GeneXpert, and Drug Susceptibility Test (DST). AFB smear, the cheapest and fastest method for detecting mycobacterium tuberculosis (Mtb), has a significant false negative rate, so that negative results do not exclude TB disease. While culture (both solid and liquid) is known as the gold standard for lab confirmation of TB disease, is much slower and because it requires sophisticated biosafety equipment to protect laboratory workers, is more expensive than AFB smear. We address the question of whether ordering and waiting on culture results is always necessary for treatment monitoring.

During treatment, a physician typically orders multiple AFB smears and at least one culture (Organization and Initiative 2010). If a physician feels patient progress through therapy is not satisfactorily, they may order additional tests (e.g., DST)

To illustrate, we selected the records of five tuberculosis patients who received treatment and their associate outcome from a longitudinal dataset containing 19,252 patients from the Republic of Moldova (SIMETB 2016). The same list of predictors as found in Kheirandish et al. (2022) is utilized to train an DNN, specifically Long-Short-Term-Memory (LSTM), classification model at a number of predetermined time points from the start of treatment. Because Mtb grow so slowly, the results of a culture test arrive in about eight weeks after the sample is taken. Although the results may be known during model training phase, they are typically not available at the actual prediction time. To assess the effect of low sensitivity of smear test on model predictions, we consider and compare two scenarios in five patients based on whether culture information is utilized or not, at prediction (Figure 4). In the first scenario, the last culture results are not utilized when predicting the treatment outcome which reflects the fact that the previous culture result has not arrived. In contrast, the second scenario involves waiting to receive culture results and utilize them to verify the smear results received earlier.

As shown in Figure 4, for Patient 1, ordering and waiting for culture results do not impact the prediction performance, as the prediction model consistently predicts the true label regardless of this additional step. Similarly, for Patient 5, the process of ordering and waiting on culture results does not help correct the wrong prediction. Consequently, ordering culture tests for both Patients 1 and 5 would merely incur additional treatment costs, and extend decision time on treatment without contributing to treatment monitoring. This holds true for Patient 2 at the 2-month follow-up, Patient 3 at the 2-month and 6-month follow-ups, and Patient 4 at 2-month follow-up.

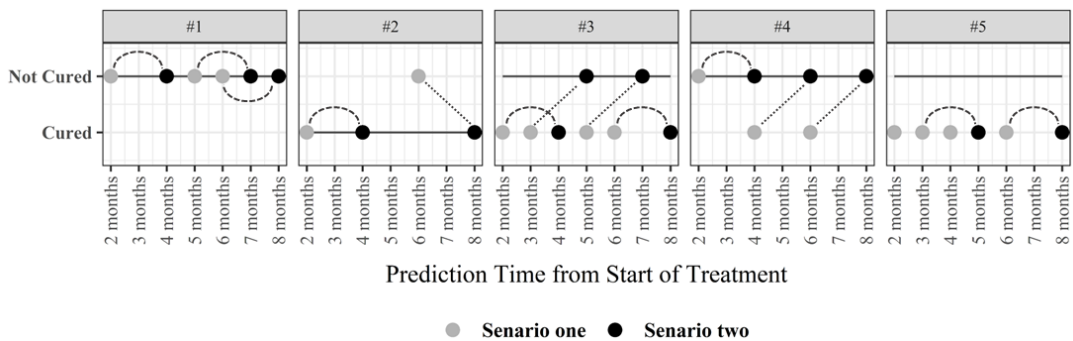

Figure 4: Treatment outcome prediction at fixed time points using all available information under two scenarios. In Scenario one, culture results have not been available but are available in Scenario two. Each box represents a patient's case, and the black solid horizontal line is the corresponding true treatment outcome. Gray and black points are predicted outcomes and the arcs link two corresponding predictions of the two scenarios. Not waiting for culture results in any of the follow-up sessions is the more advantageous decision for Patients 1 and 5.

Discovering the situations in which ordering smear tests is sufficient, despite its low sensitivity, has the potential to improve the efficacy of treatment monitoring while saving time and medical resources. Examining whether the impact of errors in smear results on prediction uncertainty is concerning can reveal these instances. This paper addresses the problem of quantifying prediction uncertainty that originates from errors in variables for Deep Neural Networks (DNN) binary classification models. Since uncertainty in smear results comes from low sensitivity of the test, in our method we assume that errors in variables follow some known discrete distribution.

*1.4 Assessing Prediction Uncertainty*

The uncertainty that comes from errors in input can be assumed as epistemic uncertainty if repeated measurements aid in reducing the error. Otherwise, it should be considered as aleatoric uncertainty since collecting more information does not help decrease the related uncertainty. However, in the computer vision literature, aleatoric is known as data uncertainty, and epistemic as model parameters uncertainty (Abdar et al. 2021). The necessity of revisiting the definitions and concepts of aleatoric and epistemic uncertainty seems imperative since deep learning models are being applied to more applications, such as modeling tabular data in addition to image modeling. An explicit distinction between these two types of uncertainty is necessary for risk-based decision making. It benefits the decision making problems by refusing or delaying the ultimate decision, such as in the learning-to-reject methods (Zhang et al. 2023), or by offering to take proper actions to reduce the uncertainty, such as in active learning (Heirung et al. 2018).

In this paper, we focus on quantifying the deep learning classification uncertainty, which manifests from the errors in variables following some discrete distribution. These errors are assumed to be aleatoric that should be represented rather than being controlled, and they originated from any sources ranging from the measurement errors to random disturbances. In Section 2, we review the existing methods of uncertainty quantification in the machine/deep learning literature. In Section 3 we introduce the mathematical formulation and solution. Finally, the results regarding applying the mathematical model on the real problem introduced in Section 1.3 will be represented and discussed in Section 5.

## 2. Literature Review

This section presents machine learning methods on representing uncertainty in classification models with a focus on the methods that distinguish aleatoric and epistemic uncertainties. The main goal of these methods is to predict the label for a set of observed predictors and the degree of uncertainty which is attached to this prediction. Bayesian inference-based methods (Theodoridis 2015) are the most common machine learning methods to quantify aleatoric and epistemic uncertainty. In these methods, a prior distribution is assumed for any desired source of epistemic uncertainty, which can be the set of parameters or any other hypothesis space. Then the knowledge about this distribution such as in Gaussian processes models (Seeger 2004) and Bayesian learning, or the set of distributions such as credal sets models (Abellán, Klir, and Moral 2006) are updated in view of additional observations. Ensemble learning, exemplified by bagging or boosting, represents another significant class of methods to quantify uncertainty. Such methods produce a set of predictors that is tempting to produce probability estimates following basic frequentist principles.

DNN models also benefit from common uncertainty quantification methods in machine learning. What makes the DNNs different from other machine learning models in essence of uncertainty quantification is that their modelling flexibility makes the epistemic uncertainty related to model hypothesis space negligible. Therefore, the main source of epistemic uncertainty is related to parameter estimation. Deep learning classification models capture aleatoric uncertainty through some transformation functions such as softmax or sigmoid layers. However, to capture epistemic uncertainty related to model parameters, Bayesian deep learning frameworks (Denker and LeCun 1990, Graves 2011, Neal 2012) were invented, in which each network weight is represented by a probability distribution. This probability distribution is updated through Bayesian deep learning. Finally, predictive distribution of labels is computed by the factorization of conditional probabilities of response given some weights and the posterior distribution of weights. Since the posterior distribution of weights cannot be obtained analytically, some approximation techniques are needed. That is how different uncertainty quantification techniques on DNN models vary from a Bayesian perspective. Laplace approximation (Mackay 1992), Markov Chain Monte Carlo (MCMC) (Neal 1992), ensemble learning (Lakshminarayanan, Pritzel, and Blundell 2017), and variational inference methods (Rossi, Michiardi, and Filippone 2019, Graves 2011, Blundell et al. 2015, Louizos and Welling 2016) are among the approximation methods to derive the posterior distribution of weights in neural networks models.

Some uncertainty quantification methods in the deep learning literature only focus on learning the total uncertainty in prediction, considering uncertainty from model parameters.

While some methods advance to distinguish between aleatoric and epistemic uncertainty of predictions. For instance, using Monte Carlo dropout (MC-dropout) as a learning technique aids in training a DNN considering uncertainty in parameters. Then, given a new query, this trained model derives the predictive distribution of labels through performing some stochastic forward passes and averaging the results (Gal and Ghahramani 2016). While this method can be used to quantify the total uncertainty of prediction, requires additional steps to distinguish between the aleatoric and epistemic components of this uncertainty.

One approach for uncertainty decomposition involves measuring both total and aleatoric uncertainty, and then calculating the epistemic uncertainty as the resulting difference. This difference equals the mutual information between the prediction and the posterior over the model parameters, which was first introduced for Gaussian Process Classifier (Houlsby et al. 2011) and was adopted as another measure of uncertainty in MC-dropout technique (Gal 2016). Depeweg et al. (2018) undertook a deliberate endeavour to measure and distinguish aleatoric and epistemic uncertainty by adopting this mutual information as the epistemic uncertainty in Bayesian deep learning with latent variables. A similar approach was recently employed to decompose aleatoric and epistemic uncertainty in DNNs with DropConnect (Mobiny et al. 2021).

Errors in variables which are also known as measurement errors, noise in data, and input uncertainty, are one of the common sources of aleatoric uncertainty in prediction models. The literature on training DNNs, whether deterministic or Bayesian, with consideration for errors in variables, is extensive when concerning regression problems (Seghouane and Fleury 2001, Van Gorp, Schoukens, and Pintelon 1998, 2000, Martin and Elster 2022). In the realm of training regression DNNs while accounting for uncertain inputs, two prevailing approaches are prominent: the utilization of DNNs with errors in variables (Martin and Elster 2022) and Bayesian Neural Networks with latent variables (Depeweg et al. 2016). These methodologies represent widely adopted strategies for handling the challenge of uncertainty in input data during the training process. However, to the best of our knowledge, comprehensive models addressing classification DNNs with uncertain inputs remain relatively unexplored. Moreover, existing DNNs that account for input uncertainty predominantly rely on an underlying assumption of normality for noise in inputs. This assumption restricts the applicability of such models primarily to prediction scenarios featuring continuous predictors.

In this paper, we propose a mathematical framework to quantify prediction uncertainty arises from errors in inputs in classification deep learning models when errors follow some discrete distribution, which complements the literature.

## 3. Mathematical Framework

In this section, a mathematical framework to quantify the uncertainty of binary classification DNN model is presented. The uncertainty is divided into the part which arises from errors in at least one of the predictors if the error follows a discrete distribution with known probability mass function. Since the notion of error is assumed to be discrete, the model is mostly practical when uncertain predictors are discrete variables. However, no specific assumptions are made regarding the notion of predictors, leaving them open to being discrete or continuous. In this methodology, during the training phase, we have access to the true values of predictors in the dataset. However, at the prediction, the true value of the new query may differ from the observed value. The primary objective of this approach is to evaluate the impact of any divergence between the true values and the observed values on the model predictions uncertainty. The model is trained based on true values, and we aim to investigate how uncertainty arises in predictions when true values deviate from what was originally observed.

In our methodology, we extend the original concept by performing calculations based on the inputs of the sigmoid layer, rather than relying on the outputs of the sigmoid layer, as initially elucidated in(Gal and Ghahramani 2016). This modification allows for a more comprehensive and refined analysis of the model uncertainty in binary classification. By considering the sigmoid layer inputs, we can obtain additional insights into the underlying probabilities and decision-making processes, leading to a more accurate assessment of the model prediction uncertainty.

Given a dataset $D = \{(\mathbf{x}_n, y_n), n = 0,1, \ldots, N\}$ where $N$ is the dataset size, each entry consists of $P$ features $\mathbf{x}_n \in \mathbb{R}^P$, and the corresponding class label $y_n \in \{0,1\}$, we consider a classification DNN model with a set of parameters $\boldsymbol{\theta}$. The neural network can have any structure but employs the sigmoid activation function and two nodes in the last layer.

We assume that $p_\xi(\mathbf{x})$ which represents the probability distribution of true values of features, $\mathbf{x}$, when features are observed as $\boldsymbol{\xi}$, is known. The predictive distribution for the target variable $y^{new}$ associated with the observed features $\boldsymbol{\xi}$ is as Equation (1).

$$p_\xi(y^{new}|D) = \sum_x \int p_\xi(y^{new}|\mathbf{x}, \boldsymbol{\theta}) p_\xi(\mathbf{x}) p(\boldsymbol{\theta}|D) \, d\boldsymbol{\theta}. \qquad \text{Eq. (1)}$$

Accepting prediction uncertainty as how spread the predictive distribution is around the mode (Gal 2016), Equation (2) derives prediction uncertainty, $U$, when $c^* = \underset{0,1}{\operatorname{argmax}} \left( P_\xi(y^{new} = c|D) \right)$. In the binary case, prediction uncertainty attains its maximum at 0.5, when $P_\xi(y^{new} = 0|D) = P_\xi(y^{new} = 1|D) = 0.5$. It attains its minimum at 0, when $P_\xi(y^{new} = c|D)=1$ for one class and $P_\xi(y^{new} = c|D) = 0$ for the other class.

$$U = 1 - P_\xi(y^{new} = c^*|D) = 1 - \sum_x \int P_\xi(y^{new} = c^*|\mathbf{x}, \boldsymbol{\theta})p_\xi(\mathbf{x})p(\boldsymbol{\theta}|D)\,d\boldsymbol{\theta}. \qquad \text{Eq. (2)}$$

Moving a layer back from the sigmoid layer, $Z_0$ and $Z_1$ are the outputs of the trained DNN model which are in fact sigmoid layer inputs. The $Z_0$ and $Z_1$ are a function of $\mathbf{x}$ and $\boldsymbol{\theta}^*$, $dnn(\mathbf{x}, \boldsymbol{\theta}^*)$, in which the function $dnn(\cdot)$ is defined by the network structure of DNN model. Since $\mathbf{x}$, and $\boldsymbol{\theta}^*$ are all random variables, it makes the $Z_0$ and $Z_1$ to be a random variable as well. Finding the probability distribution of $Z_0$ and $Z_1$ is not trivial due to the nonlinearity of the function, $dnn(\cdot)$. However, the mean and variance of the $Z_0$ and $Z_1$ can be estimated.

When $z_0$ and $z_1$ are predicted by the DNN model, $p_\xi(y^{new})$ is estimated by $\hat{P}_\xi(y^{new} = 0) = \sigma_0(z_0, z_1) = \frac{e^{-z_0}}{e^{-z_0}+e^{-z_1}}$ and $\hat{P}_\xi(y^{new} = 1) = \sigma_1(z_0, z_1) = \frac{e^{-z_1}}{e^{-z_0}+e^{-z_1}}$. Without compromising the generality of method, we derive expected value of $\hat{P}_\xi(y^{new} = 1)$ assuming that $c^* = \underset{0,1}{\operatorname{argmax}}\left(P_\xi(y^{new} = c|D)\right) = 1$ We start with quadratic Taylor approximation of the sigmoid function $\sigma_1(Z_0, Z_1)$ around $\mu_0 = E[Z_0]$ and $\mu_1 = E[Z_1]$ as follows:

$$\begin{aligned}\sigma_1(Z_0, Z_1) &\approx Q(Z_0, Z_1) \\ &= \sigma_1(\mu_0, \mu_1) + \sigma_1'(\mu_0, \mu_1)(Z_0 - \mu_0) + \sigma_1'(\mu_0, \mu_1)(Z_1 - \mu_1) \\ &\quad + \frac{1}{2}\sigma_1''(\mu_0, \mu_1)(Z_0 - \mu_0)^2 + \frac{1}{2}\sigma_1''(\mu_0, \mu_1)(Z_1 - \mu_1)^2 \\ &\quad + \sigma_1''(\mu_0, \mu_1)(Z_0 - \mu_0)(Z_1 - \mu_1).\end{aligned} \qquad \text{Eq. (3)}$$

Taking derivatives of the sigmoid functions $\sigma_0$ and $\sigma_1$, i.e.,

$$\frac{\partial \sigma_i(z_0, z_1)}{\partial z_i} = \sigma_i(z_0, z_1)[1 - \sigma_i(z_0, z_1)], i = 0,1;$$

$$\frac{\partial \sigma_i(z_0, z_1)}{\partial z_j} = -\sigma_i(z_0, z_1)\sigma_j(z_0, z_1), i = 0, 1; j = 0, 1, j \neq i,$$
Eq. (4)

$Q(Z_0, Z_1)$ can be rewritten as Equation (5).

$$Q(Z_0, Z_1) = \sigma_1(\mu_0, \mu_1) + \sigma_1(\mu_0, \mu_1)\sigma_0(\mu_0, \mu_1)[Z_1 - Z_0 - (\mu_1 - \mu_0)]$$

$$+ \frac{1}{2}\sigma_0(\mu_0, \mu_1)\sigma_1(\mu_0, \mu_1)[\sigma_0(\mu_0, \mu_1) - \sigma_1(\mu_0, \mu_1)][(Z_1 - \mu_1)^2$$

$$+ (Z_0 - \mu_0)^2 - 2(Z_1 - \mu_1)(Z_0 - \mu_0)].$$
Eq. (5)

Defining $\Delta = Z_1 - Z_0$ and $\mu_\Delta = \mu_1 - \mu_0$, Equation (5) is rewritten as

$$Q(Z_0, Z_1) = \sigma_1(\mu_0, \mu_1)$$

$$+ \sigma_1(\mu_0, \mu_1)\sigma_0(\mu_0, \mu_1)[\Delta - \mu_\Delta]$$

$$+ \frac{1}{2}\sigma_0(\mu_0, \mu_1)\sigma_1(\mu_0, \mu_1)[\sigma_0(\mu_0, \mu_1) - \sigma_1(\mu_0, \mu_1)](\Delta - \mu_\Delta)^2.$$
Eq. (6)

By taking the expected value from both sides of Equation (6) and using the fact that $E[\Delta - \mu_\Delta] = 0$ and $\sigma_0(\mu_0, \mu_1) = 1 - \sigma_1(\mu_0, \mu_1)$, we derive Equation (7).

$$E[Q(Z_0, Z_1)] = \sigma_1(\mu_0, \mu_1)$$

$$+ \frac{1}{2}\sigma_1(\mu_0, \mu_1)[1 - \sigma_1(\mu_0, \mu_1)][1 - 2\sigma_1(\mu_0, \mu_1)]Var[\Delta].$$
Eq. (7)

To derive $\mu_0$, $\mu_1$, and $Var[\Delta]$, it is important to note that the randomness of $Z_i$ in equation (6) arises from both parameters ($\boldsymbol{\theta}$) and error in predictors ($\mathbf{x}$). Let $z_i$ denote the conditional expectation $E[Z_i|\mathbf{x}, \boldsymbol{\theta}]$, then,

$$\mu_i = \mathrm{E}[Z_i] = \sum_{\mathbf{x}} \int z_i p_\xi(\mathbf{x}) p(\boldsymbol{\theta}|D)\, d\boldsymbol{\theta}\,, i = 0,1. \qquad \text{Eq. (8)}$$

$$Var[\Delta] = E[\Delta^2] - (\mathrm{E}[\Delta])^2$$

$$= \sum_{\mathbf{x}} \int (z_1 - z_0)^2 p_\xi(\mathbf{x}) p(\boldsymbol{\theta}|D)\, d\boldsymbol{\theta}$$

$$-\left(\sum_{\mathbf{x}} \int (z_1 - z_0) p_\xi(\mathbf{x}) p(\boldsymbol{\theta}|\mathbf{x})\, d\boldsymbol{\theta}\right)^2. \qquad \text{Eq. (9)}$$

The posterior distribution of parameters which is denoted as $p(\boldsymbol{\theta}|\mathbf{D})$ in equations (1), (8), and (9) is intractable (Hüllermeier and Waegeman 2021). Therefore, there is a need for an approximation technique to calculate the integral in these equations. Our method is flexible to use various uncertainty quantification techniques for DNN models which are available in the literature, such as Bayesian neural networks (Denker and LeCun 1990, Graves 2011, Neal 2012), MC-dropout (Gal and Ghahramani 2016), variational inference (Rossi, Michiardi, and Filippone 2019, Graves 2011, Blundell et al. 2015, Louizos and Welling 2016), and ensemble learning (Lakshminarayanan, Pritzel, and Blundell 2017). To illustrate, we adopt an ensemble learning model in the training phase. The model obtains $T$ bootstrap samples from the trainset, each generating a set of trained parameters. The $\mu_i$ and $Var[\Delta]$ in Equations (8) and (9) are approximated by $\hat{\mu}_i$ and $\hat{Var}[\Delta]$ in Equations (10) and (11), respectively.

$$\hat{\mu}_i = \frac{1}{T} \sum_{\mathbf{x}} \sum_{t=1}^{T} z_i^t p_\xi(\mathbf{x})\,, i = 0,1. \qquad \text{Eq. (10)}$$

$$\hat{Var}[\Delta] = \frac{1}{T} \sum_{\mathbf{x}} \sum_{t=1}^{T} (z_1^t - z_0^t)^2 p_\xi(\mathbf{x}) - \left(\frac{1}{T} \sum_{\mathbf{x}} \sum_{t=1}^{T} (z_1^t - z_0^t) p_\xi(\mathbf{x})\right)^2 \qquad \text{Eq. (11)}$$

Lastly, Prediction uncertainty in Equation (2) can be estimated by Equation (11). Since the prediction uncertainty is estimated considering the uncertainty in parameters and errors in predictors, it is specifically referred to as Errors in Variables (EIV) uncertainty, $\widehat{U}_{EIV}$

$$\widehat{U}_{EIV} = 1 - \sigma_1(\hat{\mu}_0, \hat{\mu}_1) - \frac{1}{2}\sigma_1(\hat{\mu}_0, \hat{\mu}_1)[1 - \sigma_1(\hat{\mu}_0, \hat{\mu}_1)][1 - 2\sigma_1(\hat{\mu}_0, \hat{\mu}_1)]V\hat{a}r[\Delta]. \quad \text{Eq. (12)}$$

The prediction uncertainty that does not originate from errors in variables can be derived when $p_\xi(\mathbf{x}) = \begin{cases} 1, & \mathbf{x} = \xi \\ 0, & \text{otherwise} \end{cases}$. Proposition 1 derived in the following indicates when the observed and true feature values for a new data point are identical, $1 - E[Q(Z_0, Z_1)]$ yields the prediction uncertainty that does not result from errors in variables but parameters. This prediction uncertainty is referred to as non-EIV uncertainty, $U_{nonEIV}$ in this research and can be derived through Equations (13-15).

$$\widehat{U}_{nonEIV} = 1 - E[Q(Z_0, Z_1)]$$

$$= 1 - \sigma_1(\hat{\mu}_0, \hat{\mu}_1) - \frac{1}{2}\sigma_1(\hat{\mu}_0, \hat{\mu}_1)[1 - \sigma_1(\hat{\mu}_0, \hat{\mu}_1)][1 - 2\sigma_1(\hat{\mu}_0, \hat{\mu}_1)]V\hat{a}r[\Delta]. \quad \text{Eq. (13)}$$

$$\hat{\mu}_i = \frac{1}{T}\sum_{t=1}^{T} z_i^t, i = 0, 1. \quad \text{Eq. (14)}$$

$$V\hat{a}r[\Delta] = \frac{1}{T}\sum_{t=1}^{T}(z_1^t - z_0^t)^2 - \left(\frac{1}{T}\sum_{t=1}^{T}(z_1^t - z_0^t)\right)^2. \quad \text{Eq. (15)}$$

**Proposition 1**: When $p_\xi(\mathbf{x}) = \begin{cases} 1, & \mathbf{x} = \xi \\ 0, & \text{otherwise} \end{cases}$, then $1 - E[Q(Z_0, Z_1)]$ derives prediction uncertainty considering the uncertainty in model parameters.

**Proof:** $E[Q(Z_0, Z_1)]$ in Equation (7) estimates predictive uncertainty in Equation (1) without any assumption about the shape of $p_\xi(\mathbf{x})$. Substituting $p_\xi(\mathbf{x})$ with $p_\xi(\mathbf{x}) = \begin{cases} 1, & \mathbf{x} = \xi \\ 0, & \text{otherwise} \end{cases}$, modifies the Equation (1) into Equation (16),

$$p_\xi(y^{new}|D) = \sum_x \int p_\xi(y^{new}|\mathbf{x}, \boldsymbol{\theta}) p_\xi(\mathbf{x}) p(\boldsymbol{\theta}|D) \, d\boldsymbol{\theta}$$

$$= \int p(y^{new}|\mathbf{x}, \boldsymbol{\theta}) p(\boldsymbol{\theta}|D) \, d\boldsymbol{\theta} \qquad \text{Eq. (16)}$$

The term $\int p(y^{new}|\mathbf{x}, \boldsymbol{\theta}) p(\boldsymbol{\theta}|D) \, d\boldsymbol{\theta}$ in Equation (16) demonstrates predictive distribution considering uncertainty in model parameters (Gal 2016, Hüllermeier and Waegeman 2021). Therefore, $1 - E[Q(Z_0, Z_1)]$ in Equation (12) estimates prediction uncertainty considering uncertainty in model parameters when $p_\xi(\mathbf{x}) = \begin{cases} 1, & \mathbf{x} = \xi \\ 0, & \text{otherwise} \end{cases}$. ∎

In classification Bayesian DNNs, the predictive distribution derived by applying an elementwise sigmoid function to the estimated expected values, $\mu_i$'s. Consequently, the quantified prediction uncertainty in these models corresponds to the first term of Equation (7), $E[\sigma_1(z_0, z_1)] = \sigma_1(\mu_0, \mu_1)$ when $p_\xi(\mathbf{x}) = \begin{cases} 1, & \mathbf{x} = \xi \\ 0, & \text{otherwise} \end{cases}$. This indicates that Bayesian DNNs quantify prediction uncertainty resulting from model parameters as a linear Taylor approximation of the sigmoid function, while our model quantifies the uncertainty through a quadratic Taylor approximation of sigmoid function.

Figures (5) and (6) respectively, provide an illustration of the proposed training and prediction algorithms. These visual representations provide a comprehensive understanding of the implemented methodology.

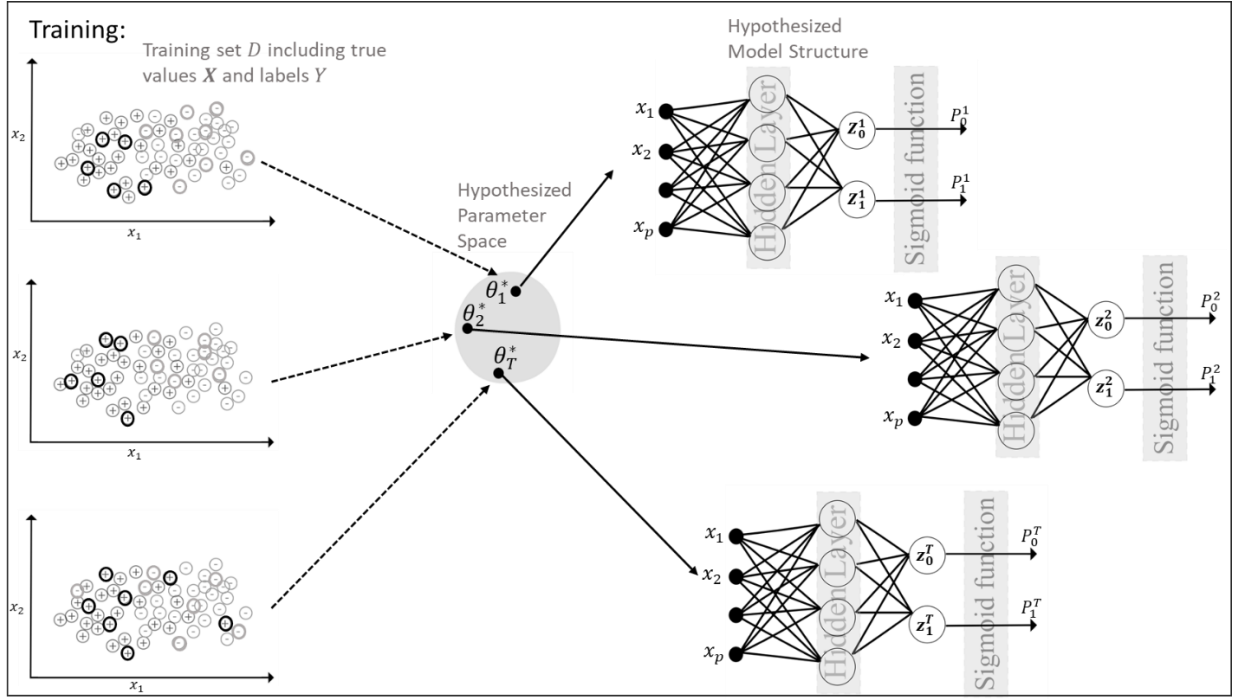

Figure 5: Proposed training algorithm to consider errors in variables and uncertainty in parameters. The train set $D$ contains $N$ points with $P$ features, and $N$ corresponding labels. It is assumed that the observed and true values of features are equal, denoted by $(x_1, x_2, ..., x_p)$. The T bootstrap samples are taken from the train set $D$, each generating a set of trained parameters, $(\theta_1^*, \theta_2^*, ..., \theta_T^*)$. Each trained model $t = 1,2, ..., T$ outputs $z_0^t$ and $z_1^t$ corresponding the labels 0 and 1. Lastly, a sigmoid function gets $z_0^t$ and $z_1^t$ as inputs to output the estimated predictive distribution as $P_0^t$ and $P_1^t$. The $P_c^t$ estimates $p(y = c|x, \theta_t^*), c = 0,1$.

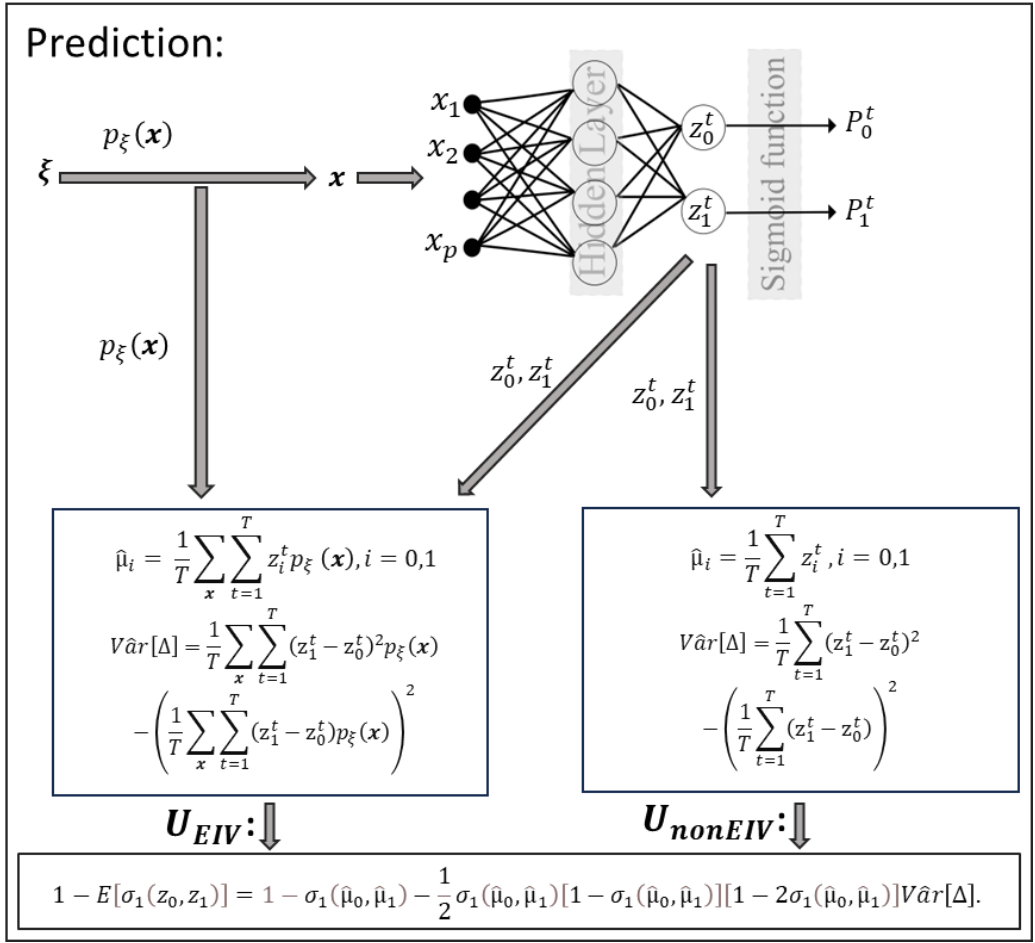

Figure 6: Proposed algorithm for assessing prediction uncertainty of DNN classification model at prediction considering errors in variables and uncertainty in parameters, denoted as EIV uncertainty, $U_{EIV}$. When $p_\xi(x) = \begin{cases} 1, & x = \xi \\ 0, & \text{otherwise} \end{cases}$, there is no error in variables and the quantified uncertainty does not arises from errors in variables but uncertainty in parameters, denoted as non-EIV uncertainty, $U_{nonEIV}$. The T bootstrap samples are taken from the train set $D$, each generating a set of trained parameters, $(\theta_1^*, \theta_2^*, \ldots, \theta_T^*)$. Each trained model $t = 1, 2, \ldots, T$ outputs $z_0^t$ and $z_1^t$ corresponding the labels 0 and 1. A sigmoid function $\sigma(\cdot)$ is used to estimate predictive distribution.

## 4. Experimental Results and Discussion

In this section, we apply the uncertainty quantification method to the tuberculosis dataset introduced in Section 1.3. The LSTM model predicts tuberculosis treatment outcome at 3, 4, 6, and 9-month follow-ups. The sensitivity and specificity of the smear test is derived from the literature (Davis et al. 2013) as 64% and 98%, respectively. The culture test is assumed to be perfect due to its high sensitivity and specificity (Caulfield and Wengenack 2016). Therefore, the results of the smear test are confirmed by culture when available for both the train and test sets. However, it is assumed that the culture results from the final time point for the test set remain pending. At each follow-up time point, the longitudinal dataset is partitioned into train and test sets (0.8/0.2) to train a LSTM model. The models are trained with 100, 200, and 500 ensembles. Within each ensemble iteration, the training set is further randomly divided into training and validation subsets. The uncertainty of a prediction is derived using the algorithm depicted in Figure 6. Greater non-EIV uncertainty indicates less model confidence on the predicted labels, which corresponds to an increased probability of encountering misclassifications.

According to Proposition 1, when all predictors are accurate, $1 - E[\sigma_1(Z_0, Z_1)]$ yields the prediction uncertainty that accounts for model parameters uncertainty and referred to as non-EIV uncertainty in this paper. Figure 7 represents a comparative analysis between the non-EIV uncertainty derived from our proposed methodology and the MC-dropout method (Gal 2016). The MC-dropout method serves as an approximation of uncertainty within Bayesian deep Gaussian models, when the epistemic uncertainty arises from model parameters. An ideal uncertainty quantification model should exhibit reduced confidence scores when misclassifications occur. This is equivalent to having a greater degree of prediction uncertainty for misclassification cases. In this essence, the ideal uncertainty quantification approach should exhibit the prediction uncertainty as 0.5 for all misclassification instances. To assess our suggested quantification method, we determine the proportion of misclassification cases in the test set with prediction uncertainty greater than several thresholds. This proportion should be equal to one for all possible thresholds for the ideal model, represented as the horizontal dashed line in Figure 7. The closer the corresponding line of a quantification method is to the horizontal dashed line, the better the model performs. Therefore, our model outperforms the MC-dropout method as it is shown in Figure 7.

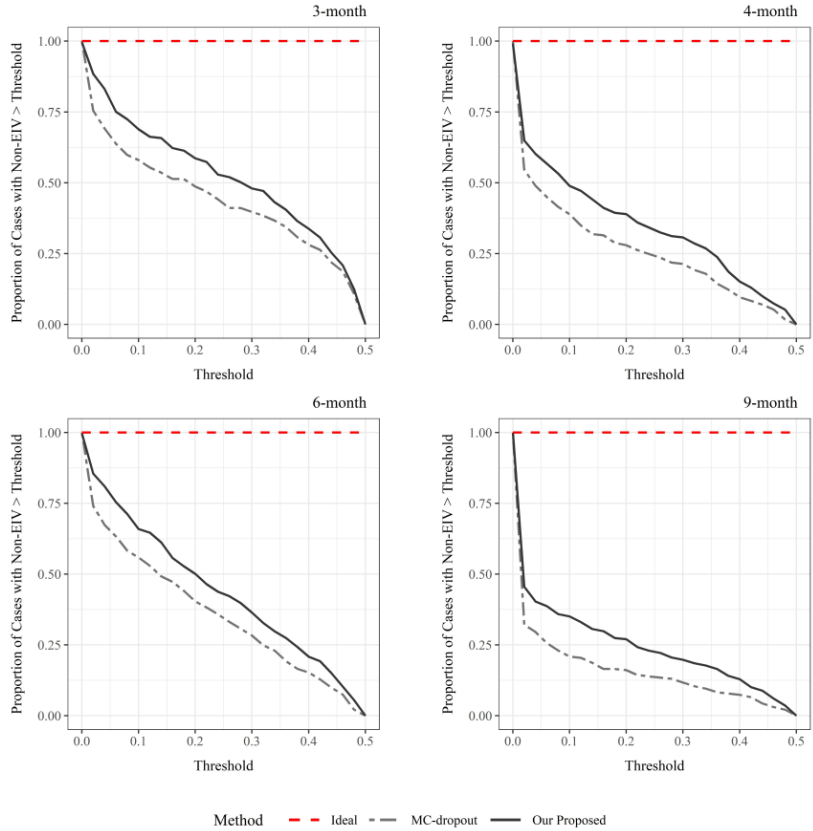

Figure 7: Comparative analysis between the non-EIV uncertainty derived from our proposed methodology and the MC-dropout method. The Y-axis indicates the proportion of misclassification cases in the test set having prediction uncertainty greater than the threshold indicated on X-axis. The dashed horizontal line corresponds to an ideal uncertainty quantification model that outputs the highest uncertainty, 0.5, for all misclassification cases.

Proposition 1 states if uncertainty in predictors reaches zero, then EIV and non-EIV uncertainties become equal. While the converse of this statement may not hold true, the condition of equal EIV and non-EIV uncertainties can be used to identify potential instances affected by errors in predictors. A graphical representation featuring EIV and non-EIV uncertainties across all data points within the test set facilitates the comprehensive evaluation of model sensitivity to errors in variables. Greater proximity of data points to the identity line signifies the sensitivity of prediction model to errors in variables. Figure 8 demonstrates that predictive models designed to predict TB treatment outcome, using the baseline and follow-up data spanning a duration of 9 months from treatment initiation are more sensitive to errors in smear results than models using information limited to 3, 4 or 6 months.

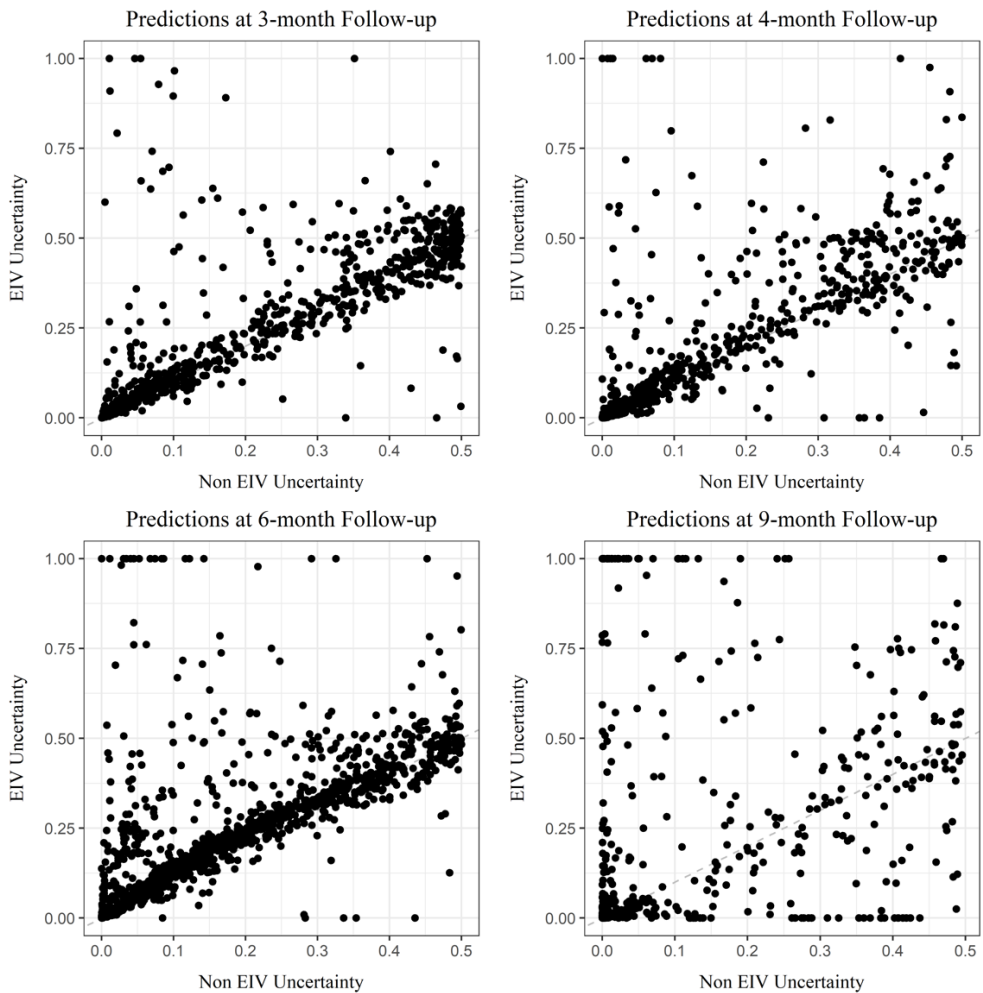

Figure 8. Quantified EIV and non-EIV uncertainties of prediction models to predict TB treatment outcome, utilizing baseline and follow-up information up to 3, 4, 6, and 9 months from start of treatment. Each point is a query in the test set. EIV and non-EIV uncertainties are referred to the predicted class for each query, which can be either cured (coded as 1) or not cured (coded as 0). The points that are close to the identity line are more probable to be insensitive to errors in smear test.

In Section 1.3, we introduced two scenarios based on whether the most recent smear results were confirmed by culture or not. For each data point within the test set, the predictive label may either change or remain consistent when culture results are considered as a confirmation of smear. A change in the predicted label signifies that the prediction uncertainty associated with the initial label has transitioned from a level below 0.5 to a level above 0.5 when the actual smear result deviates from the observed one. This indicates that errors in smear results affect model predictions for that specific data point.

Figure 9 highlights that EIV and non-EIV uncertainties when data points in the test set are categorized based on whether their predicted label changes or not, when the observed and true smear values differ. It demonstrates that when predicted labels are not prone to alteration in the presence of smear errors, EIV and non-EIV uncertainties tend to closely align. Additionally, it underscores the need to accept a certain degree of risk when determining the sensitivity of predictions to smear errors based on the proximity of EIV and non-EIV uncertainties. It should also be noted that although substantial differences between EIV and non-EIV uncertainties can pinpoint cases in which predictions are susceptible to change, they do not reveal the specific direction of these changes. Alterations in predicted labels can either transform misclassified labels into accurate ones or vice versa. The risk of encountering misclassifications should be placed solely based on non-EIV uncertainty.

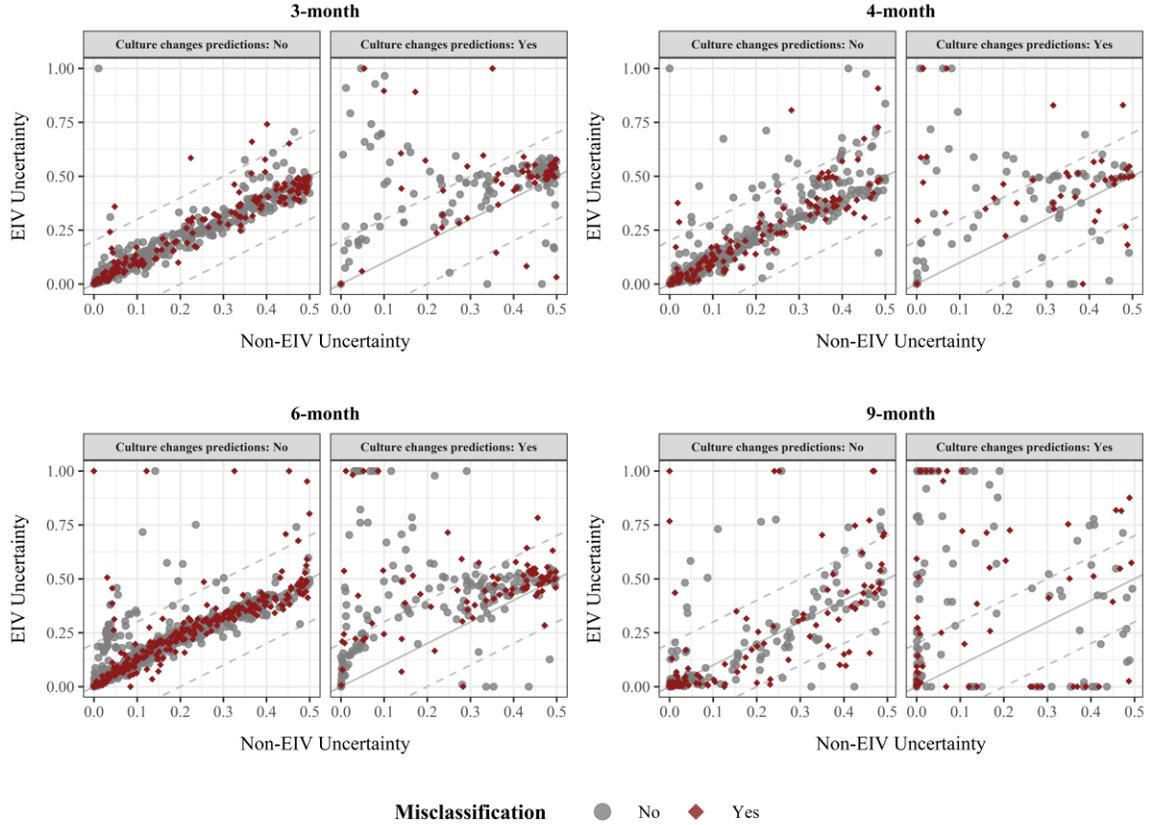

Figure 9. Queries for which the predicted label will alter with errors in smear readings, deviating from the identity line. Prediction alterations may occur for both misclassified and correctly classified cases. The solid line represents the identity line where the EIV and non-EIV uncertainty are equal. The two dashed lines depict a 0.2-unit difference between EIV and non-EIV uncertainties.

## 5. Conclusion

In this article, we present a mathematical framework designed for quantifying prediction uncertainty in deep neural network models when input variables contain error. Under the assumption that these errors follow a known discrete distribution, our model can assess the effect of these errors on predictions.

The model outputs prediction uncertainty considering errors in both variables and model parameters, referred to EIV uncertainty in this article. Additionally, the model provides a counterpart measure referred as non-EIV uncertainty, which disregards errors in variables. It is proved that when all predictors are accurate, EIV and non-EIV uncertainties are equal. This is used to identify a way to detect the instances whose predictions are sensitive to errors in variables. We apply our proposed framework to a dataset containing longitudinal information of tuberculosis patients. Four distinct deep neural network models are trained to predict binary tuberculosis treatment outcome at 3, 4, 6, and 9 months from treatment initiation, accounting for errors associated with smear results.

The results demonstrate that the non-EIV uncertainty derived from our model outperforms the MC-dropout method, which primarily accounts for uncertainty in model parameters. It is worth noting that MC-dropout is a widely utilized technique for estimating Bayesian deep Gaussian models but cannot be use to assess EIV with discrete predictors. Our results show that proximity between EIV and non-EIV uncertainties can help identify instances robust to prediction alterations in the presence of errors in predictors. This involves accepting some level of risk dependent on the selected proximity threshold. Consequently, one promising future work is to further investigate the behavior of EIV and non-EIV uncertainty. This helps establish effective metrics and associated thresholds for determining when errors in predictors pose substantial challenges.

# References

Abdar, Moloud, Farhad Pourpanah, Sadiq Hussain, Dana Rezazadegan, Li Liu, Mohammad Ghavamzadeh, Paul Fieguth, Xiaochun Cao, Abbas Khosravi, and U Rajendra Acharya. 2021. "A review of uncertainty quantification in deep learning: Techniques, applications and challenges." *Information fusion* 76:243-297.

Abellán, Joaquín, George J Klir, and Serafín Moral. 2006. "Disaggregated total uncertainty measure for credal sets." *International Journal of General Systems* 35 (1):29-44.

Blundell, Charles, Julien Cornebise, Koray Kavukcuoglu, and Daan Wierstra. 2015. "Weight uncertainty in neural network." International conference on machine learning.

Caulfield, Adam J, and Nancy L Wengenack. 2016. "Diagnosis of active tuberculosis disease: From microscopy to molecular techniques." *Journal of Clinical Tuberculosis and Other Mycobacterial Diseases* 4:33-43.

Chang, An-Hsing, Li-Kai Yang, Rua-Huan Tsaih, and Shih-Kuei Lin. 2022. "Machine learning and artificial neural networks to construct P2P lending credit-scoring model: A case using Lending Club data." *Quantitative Finance and Economics* 6 (2):303-325.

Davis, J Lucian, Adithya Cattamanchi, Luis E Cuevas, Philip C Hopewell, and Karen R Steingart. 2013. "Diagnostic accuracy of same-day microscopy versus standard microscopy for pulmonary tuberculosis: a systematic review and meta-analysis." *The Lancet infectious diseases* 13 (2):147-154.

De Angeli, Kevin, Shang Gao, Andrew Blanchard, Eric B Durbin, Xiao-Cheng Wu, Antoinette Stroup, Jennifer Doherty, Stephen M Schwartz, Charles Wiggins, and Linda Coyle. 2022. "Using ensembles and distillation to optimize the deployment of deep learning models for the classification of electronic cancer pathology reports." *JAMIA open* 5 (3):ooac075.

Denker, John, and Yann LeCun. 1990. "Transforming neural-net output levels to probability distributions." *Advances in neural information processing systems* 3.

Depeweg, Stefan, Jose-Miguel Hernandez-Lobato, Finale Doshi-Velez, and Steffen Udluft. 2018. "Decomposition of uncertainty in Bayesian deep learning for efficient and risk-sensitive learning." International Conference on Machine Learning.

Depeweg, Stefan, José Miguel Hernández-Lobato, Finale Doshi-Velez, and Steffen Udluft. 2016. "Learning and policy search in stochastic dynamical systems with bayesian neural networks." *arXiv preprint arXiv:1605.07127*.

Der Kiureghian, Armen, and Ove Ditlevsen. 2009. "Aleatory or epistemic? Does it matter?" *Structural safety* 31 (2):105-112.

Gal, Yarin. 2016. Uncertainty in deep learning. PhD thesis, University of Cambridge.

Gal, Yarin, and Zoubin Ghahramani. 2016. "Dropout as a bayesian approximation: Representing model uncertainty in deep learning." international conference on machine learning.

Graves, Alex. 2011. "Practical variational inference for neural networks." *Advances in neural information processing systems* 24.

Heirung, Tor Aksel N, Joel A Paulson, Shinje Lee, and Ali Mesbah. 2018. "Model predictive control with active learning under model uncertainty: Why, when, and how." *AIChE Journal* 64 (8):3071-3081.

Hora, Stephen C. 1996. "Aleatory and epistemic uncertainty in probability elicitation with an example from hazardous waste management." *Reliability Engineering & System Safety* 54 (2-3):217-223.

Houlsby, Neil, Ferenc Huszár, Zoubin Ghahramani, and Máté Lengyel. 2011. "Bayesian active learning for classification and preference learning." *arXiv preprint arXiv:1112.5745*.

Hüllermeier, Eyke, and Willem Waegeman. 2021. "Aleatoric and epistemic uncertainty in machine learning: An introduction to concepts and methods." *Machine Learning* 110:457-506.

Kheirandish, Maryam, Donald Catanzaro, Valeriu Crudu, and Shengfan Zhang. 2022. "Integrating landmark modeling framework and machine learning algorithms for dynamic prediction of tuberculosis treatment outcomes." *Journal of the American Medical Informatics Association* 29 (5):900-908.

Lakshminarayanan, Balaji, Alexander Pritzel, and Charles Blundell. 2017. "Simple and scalable predictive uncertainty estimation using deep ensembles." *Advances in neural information processing systems* 30.

Louizos, Christos, and Max Welling. 2016. "Structured and efficient variational deep learning with matrix gaussian posteriors." International conference on machine learning.

Mackay, David John Cameron. 1992. *Bayesian methods for adaptive models*: California Institute of Technology.

Martin, J, and C Elster. 2022. "Aleatoric Uncertainty for Errors-in-Variables Models in Deep Regression." *Neural Processing Letters*:1-20.

Mobiny, Aryan, Pengyu Yuan, Supratik K Moulik, Naveen Garg, Carol C Wu, and Hien Van Nguyen. 2021. "Dropconnect is effective in modeling uncertainty of bayesian deep networks." *Scientific reports* 11 (1):5458.

Mourtas, Spyridon D, Vasilios N Katsikis, Emmanouil Drakonakis, and Stelios Kotsios. 2023. "Stabilization of stochastic exchange rate dynamics under central bank intervention using neuronets." *International Journal of Information Technology & Decision Making* 22 (02):855-883.

Neal, Radford. 1992. "Bayesian learning via stochastic dynamics." *Advances in neural information processing systems* 5.

Neal, Radford M. 2012. *Bayesian learning for neural networks*. Vol. 118: Springer Science & Business Media.

Organization, World Health, and Stop TB Initiative. 2010. *Treatment of tuberculosis: guidelines*: World Health Organization.

Rossi, Simone, Pietro Michiardi, and Maurizio Filippone. 2019. "Good initializations of variational bayes for deep models." International Conference on Machine Learning.

Seeger, Matthias. 2004. "Gaussian processes for machine learning." *International journal of neural systems* 14 (02):69-106.

Seghouane, A-K, and Gilles Fleury. 2001. "A cost function for learning feedforward neural networks subject to noisy inputs." Proceedings of the sixth international symposium on signal processing and its applications (Cat. No. 01EX467).

Seo, Jigu, and Sungwook Park. 2023. "Optimizing model parameters of artificial neural networks to predict vehicle emissions." *Atmospheric Environment* 294:119508.

SIMETB. 2016. edited by SIMETB.

Theodoridis, Sergios. 2015. *Machine learning: a Bayesian and optimization perspective*: Academic press.

Van Gorp, Jürgen, Johan Schoukens, and Rik Pintelon. 1998. "The errors-in-variables cost function for learning neural networks with noisy inputs."  *Intelligent Engineering through Artificial Neural Networks* 8:141-146.

Van Gorp, Jürgen, Johan Schoukens, and Rik Pintelon. 2000. "Learning neural networks with noisy inputs using the errors-in-variables approach."  *IEEE Transactions on Neural Networks* 11 (2):402-414.

Walker, Warren E, Poul Harremoës, Jan Rotmans, Jeroen P Van Der Sluijs, Marjolein BA Van Asselt, Peter Janssen, and Martin P Krayer von Krauss. 2003. "Defining uncertainty: a conceptual basis for uncertainty management in model-based decision support."  *Integrated assessment* 4 (1):5-17.

Zhang, Xu-Yao, Guo-Sen Xie, Xiuli Li, Tao Mei, and Cheng-Lin Liu. 2023. "A survey on learning to reject."  *Proceedings of the IEEE* 111 (2):185-215.